\documentclass[runningheads]{llncs}
\usepackage{graphicx}
\usepackage{amsmath,amssymb} 
\usepackage{color}

\usepackage{epsfig}
\usepackage{bm}
\usepackage{booktabs}
\usepackage{multirow}
\usepackage{cite}
\usepackage{mathtools}
\usepackage{enumitem}
\usepackage{bbm}
\newcommand{\xhdr}[1]{\vspace{3pt}\noindent \textbf{#1}}
\newcommand{\etal}{\textit{et al}.}
\newcommand{\ie}{\textit{i}.\textit{e}.}

\newcommand{\jianwei}[1]{#1}

\begin{document}
\title{Graph R-CNN for Scene Graph Generation} 

\titlerunning{Graph R-CNN}
%
\author{Jianwei Yang\inst{1}\thanks{Equal contribution} \and
Jiasen Lu\inst{1}$^{\star}$ \and
Stefan Lee\inst{1} \and
Dhruv Batra\inst{1,2} \and
Devi Parikh\inst{1,2}
}
%
\authorrunning{Yang and Lu \etal}
%

\institute{$^1$Georgia Institute of Technology \quad $^2$Facebook AI Research \\
	\email{ \tt \small \{jw2yang, jiasenlu, steflee, dbatra, parikh\}@gatech.edu}}
\maketitle              
\begin{abstract}
We propose a novel scene graph generation model called Graph R-CNN, that is both effective and efficient at detecting objects and their relations in images. Our model contains a Relation Proposal Network (RePN) that efficiently deals with the quadratic number of potential relations between objects in an image. We also propose an attentional Graph Convolutional Network (aGCN) that effectively captures contextual information between objects and relations. Finally, we introduce a new evaluation metric that is more holistic and realistic than existing metrics. We report state-of-the-art performance on scene graph generation as evaluated using both existing and our proposed metrics.

\end{abstract}

\section{Introduction}
Visual scene understanding has traditionally focused on identifying \emph{objects in images} -- learning to predict their presence (\ie~image classification \cite{krizhevsky2012imagenet, szegedy2015going, he2016deep}) and spatial extent (\ie~object detection \cite{girshick2014rich, redmon2016you, liu2016ssd} or segmentation \cite{lin2017feature}). These object-centric techniques have matured significantly in recent years, however, representing scenes as collections of objects fails to capture relationships which may be essential for scene understanding.

\begin{figure}[t]
  \begin{center}
    \includegraphics[width=1\textwidth, trim=0 0.5cm 0 1cm, clip]{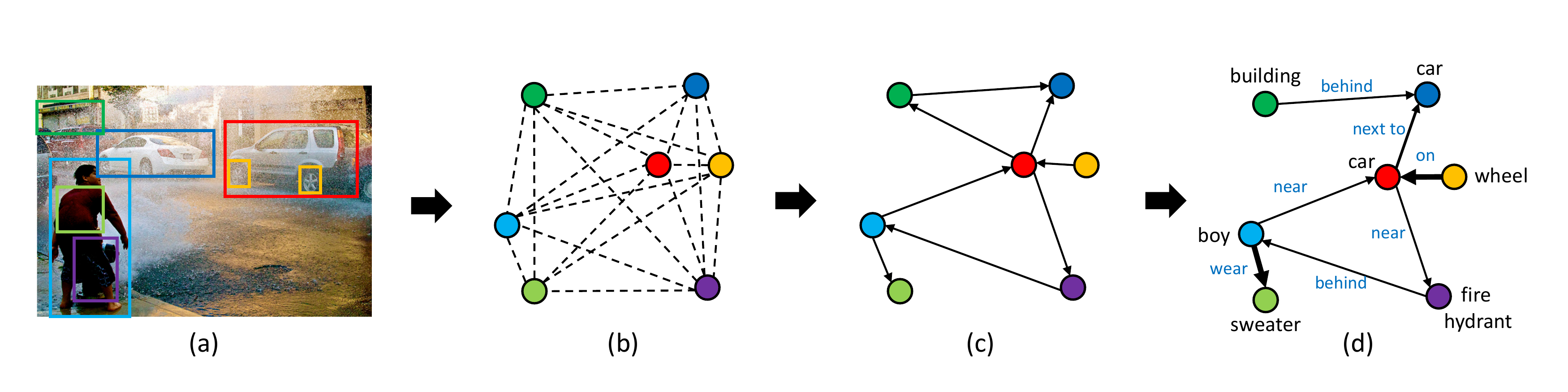}
  \end{center}
\caption{Given an image (a), our proposed approach first extracts a set of objects visible in the scene and considers possible relationships between all nodes (b). Then it prunes unlikely relationships using a learned measure of `relatedness', producing a sparser candidate graph structure (c). Finally, an attentional graph convolution network is applied to integrate global context and update object node and relationship edge labels.}
\label{fig:teaser}
\end{figure}

A recent work \cite{Johnson2015CVPR} has instead proposed representing visual scenes as graphs containing objects, their attributes, and the relationships between them. These \emph{scene graphs} form an interpretable structured representation of the image that can support higher-level visual intelligence tasks such as captioning \cite{wu2017image, Lu2018Neural}, visual question answering \cite{antol2015vqa, teney2016graph, wu2017image, wang2017vqa, wang2017fvqa, johnson2017clevr}, and image-grounded dialog \cite{das2016visual}. While scene graph representations hold tremendous promise, extracting scene graphs from images -- efficiently and accurately -- is challenging. The natural approach of considering every pair of nodes (objects) as a potential edge (relationship) -- essentially reasoning over fully-connected graphs -- is often effective in modeling contextual relationships but scales poorly (quadratically) with the number of objects, quickly becoming impractical. 
The naive fix of randomly sub-sampling edges to be considered is more efficient but not as effective since the distribution of interactions between objects is far from random -- take Fig.~\ref{fig:teaser}(a) as an example, it is much more likely for a `car' and `wheel' to have a relationship than a `wheel' and `building'. Furthermore, the types of relationships that typically occur between objects are also highly dependent on those objects. 

\xhdr{Graph R-CNN.} In this work, we propose a new framework, Graph R-CNN, for scene graph generation which effectively leverages object-relationship regularities through two mechanisms to intelligently sparsify and reason over candidate scene graphs. Our model can be factorized into three logical stages: 1) object node extraction, 2) relationship edge pruning, and 3) graph context integration, which are depicted in Fig.~\ref{fig:teaser}.
In the object node extraction stage, we utilize a standard object detection pipeline \cite{ren2015faster}. This results in a set of localized object regions as shown in Fig.~\ref{fig:teaser}b.
We introduce two important novelties in the rest of the pipeline to incorporate the real-world regularities in object relationships discussed above.
First, we introduce a relation proposal network (RePN) that learns to efficiently compute \emph{relatedness scores} between object pairs which are used to intelligently prune unlikely scene graph connections (as opposed to random pruning in prior work). A sparse post-pruning graph is shown in Fig.~\ref{fig:teaser}c.
Second, given the resulting sparsely connected scene graph candidate, we apply an attentional graph convolution network (aGCN) to propagate higher-order context throughout the graph -- updating each object and relationship representation based on its neighbors. In contrast to existing work, we predict per-node edge attentions, enabling our approach to learn to modulate information flow across unreliable or unlikely edges. We show refined graph labels and edge attentions (proportional to edge width) in Fig.~\ref{fig:teaser}d.

To validate our approach, we compare our performance with existing methods on the Visual Genome \cite{krishna2017visual} dataset and find that our approach achieves an absolute gain of 5.0 on Recall@50 for scene graph generation \cite{xu2017scene}. We also perform extensive model ablations and quantify the impact of our modeling choices.

\xhdr{Evaluating Scene Graph Generation.} %
Existing metrics for scene graph generation are based on recall of 
$\langle$subject, predicate, object$\rangle$ triplets (e.g.~\texttt{SGGen} from \cite{krishna2017visual}) 
or of objects and predicates given ground truth object localizations (e.g.~\texttt{PredCls} and \texttt{PhrCls} from \cite{krishna2017visual}). 
In order to expose a problem with these metrics, consider a method that mistakes the boy in Fig.~\ref{fig:teaser}a as a man but 
otherwise identifies that he is 1) standing behind a fire hydrant, 
2) near a car, and 
3) wearing a sweater. Under the triplet-based metrics, 
this minor error (boy vs man) would be heavily penalized despite most of the boy's relationships being correctly identified. 
Metrics that provide ground-truth regions side-step this problem by focusing strictly on relationship prediction 
but cannot accurately reflect the test-time performance of the entire scene graph generation system. 

To address this mismatch, we introduce a novel evaluation metric (\texttt{SGGen+}) that more holistically evaluates 
the performance of scene graph generation with respect to objects, attributes (if any), and relationships. 
Our proposed metric \texttt{SGGen+} computes the total recall for singleton entities (objects and predicates), 
pair entries $\langle$object, attribute$\rangle$ (if any), and triplet entities $\langle$subject, predicate, object$\rangle$. 
We report results on existing methods under this new metric and find our approach also outperforms the state-of-the-art significantly. 
More importantly, this new metric provides a more robust and holistic measure of similarity between generated and ground-truth scene graphs.

\xhdr{Summary of Contributions.} Concretely, this work addresses the scene graph generation problem by introducing 
a novel model (Graph R-CNN), which can leverage object-relationship regularities, 
and proposes a more holistic evaluation metric (\texttt{SGGen+}) 
for scene graph generation. 
We benchmark our model against existing approaches on standard metrics and this new measure -- outperforming existing approaches.
\section{Related Work}

\xhdr{Contextual Reasoning and Scene Graphs.}
The idea of using context to improve scene understanding has a long history in computer vision \cite{parikh2008appearance, oliva2007role, ladicky2010graph, rabinovich2007objects}. More recently, inspired by representations studied by the graphics community, Johnson \etal \cite{Johnson2015CVPR} introduced the problem of extracting scene graphs from images, which generalizes the task of object detection \cite{girshick2014rich, girshick2015fast, ren2015faster, redmon2016you, liu2016ssd} to also detecting relationships and attributes of objects.

\xhdr{Scene Graph Generation}. A number of approaches have been proposed for the detection of both objects and their relationships \cite{lu2016visual, zhuang2017towards, peyre2017weakly, zhang2017ppr, zhang2017visual, xu2017scene, li2017vip, liang2017deep, dai2017detecting, li2017scene, newell2017pixels, zellers2017neural}. Though most of these works point out that reasoning over a quadratic number of relationships in the scene graph is intractable, each resorted to heuristic methods like random sampling to address this problem. Our work is the first to introduce a trainable relationship proposal network (RePN) that learns to prune unlikely relationship edges from the graph without sacrificing efficacy. RePN provides high-quality relationship candidates, which we find improves overall scene graph generation performance. 

Most scene graph generation methods also include some mechanisms for context propagation and reasoning over a candidate scene graph in order to refine the final labeling. 
In \cite{xu2017scene}, Xu \etal decomposed the problem into two sub-graphs -- one for objects and one for relationships -- and performed message passing. 
Similarly, in \cite{li2017vip}, the authors propose two message-passing strategies (parallel and sequential) for propagating information between objects and relationships. 
Dai \etal \cite{dai2017detecting} address model the scene graph generation process as inference on a conditional random field (CRF). 
Newell \etal ~\cite{newell2017pixels} proposed to directly generate scene graphs  
from image pixels without the use of object detector based on associative graph embeddings. 
In our work, we develop a novel attentional graph convolutional network (aGCN) to update node and relationship 
representations by propagating context between nodes in candidate scene graphs -- operating both on visual and semantic features. 
While similar in function to the message-passing based approach above, aGCN is highly efficient and can learn to place 
attention on reliable edges and dampen the influence of unlikely ones. 

A number of previous approaches have noted the strong regularities in scene graph generation which motivate our approach.
In \cite{lu2016visual}, Lu \etal ~integrates semantic priors from language to improve the detection of meaningful relationships between objects. Likewise, Li \etal ~\cite{li2017scene} demonstrated that region captions can also provide useful context for scene graph generation. Most related to our motivation, Zeller \etal \cite{zellers2017neural} formalize the notion of motifs (\ie, regularly occurring graph structures) and examine their prevalence in the Visual Genome dataset \cite{krishna2017visual}. The authors also propose a surprisingly strong baseline which directly uses frequency priors to predict relationships -- explicitly integrating regularities in the graph structure. 

\xhdr{Relationship Proposals}. Our Relationship Proposal Network  (RePN) is inspired and relates strongly to the region proposal network (RPN) of faster R-CNN \cite{ren2015faster} used in object detection. Our RePN is also similar in spirit to the recently-proposed relationship proposal network (Rel-PN) \cite{zhang2017relationship}. There are a number of subtle differences between these approaches. The Rel-PN model independently predicts proposals for subject, objects and predicates, and then re-scores all valid triples, while our RePN generates relations conditioned on objects, allowing it to learn object-pair relationship biases. Moreover, their approach is class agnostic and has not been used for scene graph generation. 

\xhdr{Graph Convolutional Networks (GCNs)}. GCNs were first proposed in \cite{kipf2016semi} in the context of semi-supervised learning. GCNs decompose complicated computation over graph data into a series of localized operations (typically only involving neighboring nodes) for each node at each time step. The structure and edge strengths are typically fixed prior to the computation. 
For completeness, we note that an upcoming publication \cite{velivckovic2017graph} has concurrently and independently developed a similar GCN attention mechanism (as aGCN) and shown its effectiveness in other (non-computer vision) contexts.
\section{Approach}
\begin{figure}[t]
\centering
\includegraphics[width=1\textwidth, trim=0 0 0 1.2cm, clip]{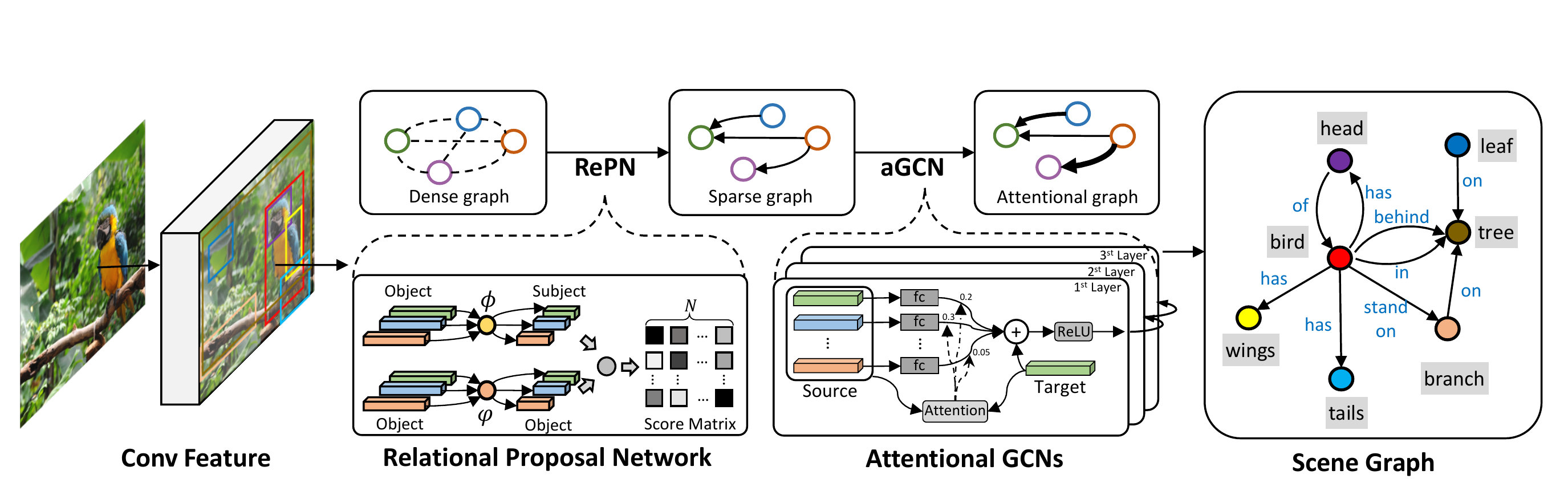}
\caption{The pipeline of our proposed Graph R-CNN framework. Given an image, our model first uses RPN to propose object regions, and then prunes the connections between object regions through our relation proposal network (RePN). Attentional GCN is then applied to integrate contextual information from neighboring nodes in the graph. Finally, the scene graph is obtained on the right side.}
\label{fig:framework}
\end{figure}

In this work, we model scene graphs as graphs consisting of image regions, relationships, and their labellings. More formally, 
let $I$ denote an image, $V$ be a set of nodes corresponding to localized object regions in $I$, $E \in {V \choose 2}$ denote the relationships (or edges) between objects, and $O$ and $R$ denote object and relationship labels respectively. Thus, the goal is to build a model for $P(S = (V, E, O, R) | I)$. In this work, we factorize the scene graph generation process into three parts:
\begin{eqnarray}
P(\mathcal{S}|\bm{I}) = \overbrace{P(\bm{V}|\bm{I})}^{\substack{\text{Object Region}\\ \text{Proposal}}}\underbrace{P(\bm{E}|\bm{V},\bm{I})}_{\substack{\text{Relationship} \\ \text{Proposal}}} \overbrace{P(\bm{R},\bm{O}|\bm{V}, \bm{E}, \bm{I})}^{\text{Graph Labeling}}
\label{Eq:SGG_Factorized}
\end{eqnarray}
which separates graph construction (nodes and edges) from graph labeling. The intuition behind this factorization is straightforward. First, the object region proposal $P(\bm{V}|\bm{I})$ is typically modeled using an off-the-shelf object detection system such as \cite{ren2015faster} to produce candidate regions. Notably, existing methods typically model the second relationship proposal term $P(\bm{E}|\bm{V},\bm{I})$ as a uniform random sampling of potential edges between vertices $\bm{V}$. In contrast, we propose a relationship proposal network (RePN) to directly model $P(\bm{E}|\bm{V},\bm{I})$ -- making our approach the first that allows for learning the entire generation process end-to-end. Finally, the graph labeling process $P(\bm{R},\bm{O}|\bm{V}, \bm{E}, \bm{I})$ is typically treated as an iterative refinement process \cite{li2017vip,xu2017scene,dai2017detecting}. A brief pipeline is shown in Fig.~\ref{fig:framework}.

In the following, we discuss the components of our proposed Graph R-CNN model corresponding to each of the terms in Eq.~\ref{Eq:SGG_Factorized}. First, we discuss our use of Faster R-CNN \cite{ren2015faster} for node generation in Section \ref{sec:rpn}. Then in Section \ref{sec:repn} we introduce our novel relation proposal network architecture to intelligently generate edges. Finally, in Section \ref{sec:agcn} we present our graph convolutional network \cite{kipf2016semi} with learned attention to adaptively integrate global context for graph labeling.

\subsection{Object Proposals}
\label{sec:rpn}
In our approach, we use the Faster R-CNN \cite{ren2015faster} framework to extract a set of $n$ object proposals from an input image. Each object proposal $i$ is associated with a spatial region $r^o_i =[x_i,y_i,w_i,h_i]$, a pooled feature vector $x^o_i$, and an initial estimated label distribution $p^o_i$ over classes $C{=}\{1,\dots,k\}$. We denote the collection of these vectors for all $n$ proposals as the matrices $R^o {\in}~\mathbb{R}^{n \times 4}$ , $X^o {\in}~ \mathbb{R}^{n \times d}$, and $P^o{\in}~\mathbb{R}^{n \times |C|}$ respectively.

\subsection{Relation Proposal Network}
\label{sec:repn}

Given the $n$ proposed object nodes from the previous step, there are $O(n^2)$ possible connections between them; however, as previously discussed, most object pairs are unlikely to have relationships due to regularities in real-world object interactions. To model these regularities, we introduce a relation proposal network (RePN) which learns to  efficiently estimate the \emph{relatedness} of an object pair.  By pruning edges corresponding to unlikely relations, the RePN can efficiently sparsify the candidate scene graph -- retaining likely edges and suppressing noise introduced from unlikely ones.

In this paper, we exploit the estimated class distributions ($P^o$) to infer relatedness -- essentially learning soft class-relationships priors. This choice aligns well with our intuition that certain classes are relatively unlikely to interact compared with some other classes. Concretely, given initial object classification distributions $P^o$, we score all $n*(n-1)$ directional pairs $\{\bm{p}^o_i, \bm{p}^o_j | i\neq j\}$, computing the relatedness as $s_{ij} = f(\bm{p}^o_i, \bm{p}^o_j) ~$ where $f(\cdot, \cdot)$ is a learned relatedness function. One straightforward implementation of $f(\cdot,\cdot)$ could be passing the concatenation $[\bm{p}^o_i, \bm{p}^o_j]$ as input to a multi-layer perceptron which outputs the score. However, this approach would consume a great deal of memory and computation given the quadratic number of object pairs. To avoid this, we instead consider an asymmetric kernel function:
\begin{equation}
\mathit{f}(\bm{p}^o_i, \bm{p}^o_j) = \langle\Phi(\bm{p}^o_i), \Psi(\bm{p}^o_j)\rangle, i \neq j
\end{equation}
where $\Phi(\cdot)$ and $\Psi(\cdot)$ are projection functions for subjects and objects in the relationships respectively\footnote{We distinguish between the first and last object in a relationship as subject and object respectively, that is, $\langle \mathtt{subject, relationship, object}\rangle$.}. This decomposition allows the score matrix $S=\{s_{ij}\}^{n \times n}$ to be computed \emph{with only two projection processes for ${X}^o$ followed by a matrix multiplication}. We use two multi-layer perceptrons (MLPs) with identical architecture (but different parameters) for $\Phi(\cdot)$ and $\Psi(\cdot)$. We also apply a sigmoid function element-wise to $S$ such that all relatedness scores range from 0 to 1.

After obtaining the score matrix for all object pairs, we sort the the scores in descending order and choose top $K$ pairs. We then apply non-maximal suppression (NMS) to filter out object pairs that have significant overlap with others. Each relationship has a pair of bounding boxes, and the combination order matters. We compute the overlap between two object pairs $\{u, v\}$ and $\{p, q\}$ as:
\begin{equation}
\small
IoU(\{u, v\}, \{p, q\}) = \frac{I({r}^o_u, {r}^o_p) + I({r}^o_v, {r}^o_q)}{U({r}^o_u, {r}^o_p) + U({r}^o_v, {r}^o_q)}
\end{equation}

\noindent where operator $I$ computes the intersection area between two boxes and $U$ the union area. The remaining $m$ object pairs are considered as candidates having meaningful relationships $\bm{E}$. With $\bm{E}$, we obtain a graph $\mathcal{G} =\left( \bm{V}, \bm{E} \right)$, which is much sparser than the original fully connected graph. Along with the edges proposed for the graph, we get the visual representations $X^r=\{\bm{x}^r_1,...,\bm{x}^r_m\}$ for all $m$ relationships by extracting features from the union box of each object pair.

\subsection{Attentional GCN}
\label{sec:agcn}
To integrate contextual information informed by the graph structure, we propose an attentional graph convolutional network (aGCN). Before we describe our proposed aGCN, let us briefly recap a `vanilla' GCN in which each node $i$ has a representation $z_i \in \mathbb{R}^d$, as proposed in \cite{kipf2016semi}. 
Briefly, for a target node $i$ in the graph, the representations of its neighboring nodes $\{z_j ~|~ j \in \mathcal{N}(i)\}$ are first transformed via a learned linear transformation $W$. Then, these transformed representations are gathered with predetermined weights $\alpha$, followed by a non-linear function $\sigma$ 
(ReLU \cite{nair2010rectified}). 
This layer-wise propagation can be written as:
\begin{equation}
\bm{z}_{i}^{(l+1)} = \sigma \left( \bm{z}_i^{(l)} + \sum_{j \in \mathcal{N}(i)} \alpha_{ij} W \bm{z}_{j}^{(l)} \right)
\label{Eq:gcn_orig}
\end{equation}
or equivalently we can collect node representations into a matrix $Z\in\mathbb{R}^{d{\times}Tn}$
\begin{equation}
\bm{z}_{i}^{(l+1)} = \sigma \left( W Z^{(l)} \bm{\alpha}_i \right)
\label{Eq:gcn_orig_sparse}
\end{equation}
for $\bm{\alpha}_i\in [0,1]^{n}$ with 0 entries for nodes not neighboring $i$ and $\bm{\alpha}_{ii}=1$. In conventional GCN, the connections in the graph are known and coefficient vector $\bm{\alpha}_i$ are preset based on the symmetrically normalized adjacency matrix of features.

In this paper, we extend the conventional GCN to an attentional version, which we refer to as aGCN, by learning to adjust $\bm{\alpha}$. To predict attention from node features, we learn a 2-layer MLP over concatenated node features and compute a softmax over the resulting scores. The the attention for node $i$ is
\begin{align}
u_{ij} &= w_{h}^T \sigma(W_{a} [\bm{z}_i^{(l)}, \bm{z}_j^{(l)}])\\
\bm{\alpha}_{i} &= \textrm{softmax}(\bm{u}_{i}), 
\label{Eq:gcn_alpha}
\end{align}
where $w_{h}$ and $W_{a}$ are learned parameters and $[\cdot,\cdot]$ is the concatenation operation. By definition, we set $\bm{\alpha}_{ii}=1$ and $\bm{\alpha}_{ij}=0 ~\forall j \notin \mathcal{N}(i)$. As attention is a function of node features, each iteration results in altered attentions which affects successive iterations.

\xhdr{aGCN for Scene Graph Generation.} Recall that from the previous sections we have a set of $N$ object regions and $m$ relationships. From these, we construct a graph $G$ with nodes corresponding to object and relationship proposals. We insert edges between relation nodes and their associated objects. We also add skip-connect edges directly between all object nodes. These connections allow information to flow directly between object nodes. Recent work has shown that reasoning about object correlation can improve detection performance \cite{hu2018relation}.
We apply aGCN to this graph to update object and relationship representations based on global context.

Note that our graph captures a number of different types of connections (\ie $\mathtt{object}\leftrightarrow\mathtt{relationship}$, $\mathtt{relationship}\leftrightarrow\mathtt{subject}$ and $ \mathtt{object}\leftrightarrow\mathtt{object}$). In addition, the information flow across each connection may be asymmetric ( the informativeness of \texttt{subject} on \texttt{relationship} might be quite different from \texttt{relationship} to \texttt{subject}). We learn different transformations for each type and ordering -- denoting the linear transform from node type $a$ to node type $b$ as $W^{ab}$ with $s{=}$subjects, $o{=}$objects, and $r{=}$relationships.
Using the same notation as in Eq.~\ref{Eq:gcn_orig_sparse} and writing object and relationship features as $Z^o$ and $Z^r$, we write the representation update for object nodes as
\begin{equation}
\bm{z}_{i}^o = \sigma ( 
\overbrace{W^{\mathtt{skip}} Z^o \bm{\alpha}^{\mathtt{skip}}}^{\substack{\text{Message from}\\ \text{Other Objects}}} + 
\overbrace{W^{sr} Z^r\bm{\alpha}^{sr} + W^{or} Z^r\bm{\alpha}^{or}}^{\substack{\text{Messages from}\\ \text{Neighboring Relationships}}}) 
\end{equation}
with $\bm{\alpha}^{\mathtt{skip}}_{ii}{=}1$ and similarly for relationship nodes as
\begin{equation}
\bm{z}_{i}^r = \sigma (\bm{z}_{i}^r + \underbrace{W^{rs} Z^{o} \bm{\alpha}^{rs} +  W^{ro} Z^{o} \bm{\alpha}^{ro}}_{\text{Messages from Neighboring Objects}}).
\end{equation}
where $\bm{\alpha}$ are computed at each iteration as in Eq.~\ref{Eq:gcn_alpha}.

One open choice is how to initialize the object and relationship node representations $z$ which could potentially be set to any intermediate feature representation or even the pre-softmax output corresponding to class labels. In practice, we run both a visual and semantic aGCN computation -- one with visual features and the other using pre-softmax outputs. In this way, we can reason about both lower-level visual details (\ie two people are likely talking if they are facing one another) as well as higher-level semantic co-occurrences (\ie cars have wheels). Further, we set the attention in the semantic aGCN to be that of the visual aGCN -- effectively modulating the flow of semantic information based on visual cues. This also enforces that real-world objects and relationships represented in both graphs interact with others in the same manner.

\subsection{Loss Function}
In Graph R-CNN, we factorize the scene graph generation process into three sub-processes: $P(\bm{R},\bm{O}|\bm{V}, \bm{E}, \bm{I})$, $P(\bm{E}|\bm{V},\bm{I})$, $P(\bm{V}|\bm{I})$, which were described above. During training, each of these sub-processes are trained with supervision. For $P(\bm{V}|\bm{I})$, we use the same loss as used in RPN, which consists of a binary cross entropy loss on proposals and a regression loss for anchors. For $P(\bm{E}|\bm{V}, \bm{I})$, we use another binary cross entropy loss on the relation proposals. For the final scene graph generation $P(\bm{R},\bm{O}|\bm{V}, \bm{E}, \bm{I})$, two multi-class cross entropy losses are used for object classification and predicate classification.
\section{Evaluating Scene Graph Generation}

Scene graph generation is naturally a structured prediction problem over attributed graphs, and how to correctly and efficiently evaluate predictions is an under-examined problem in prior work on scene graph generation. We note that graph similarity based on minimum graph edit distance has been well-studied in graph theory \cite{gao2010survey}; however, computing exact solution is NP-complete and approximation APX-hard \cite{lin1994hardness}.

\begin{figure}[t]
  \begin{center}
    \includegraphics[width=1\textwidth]{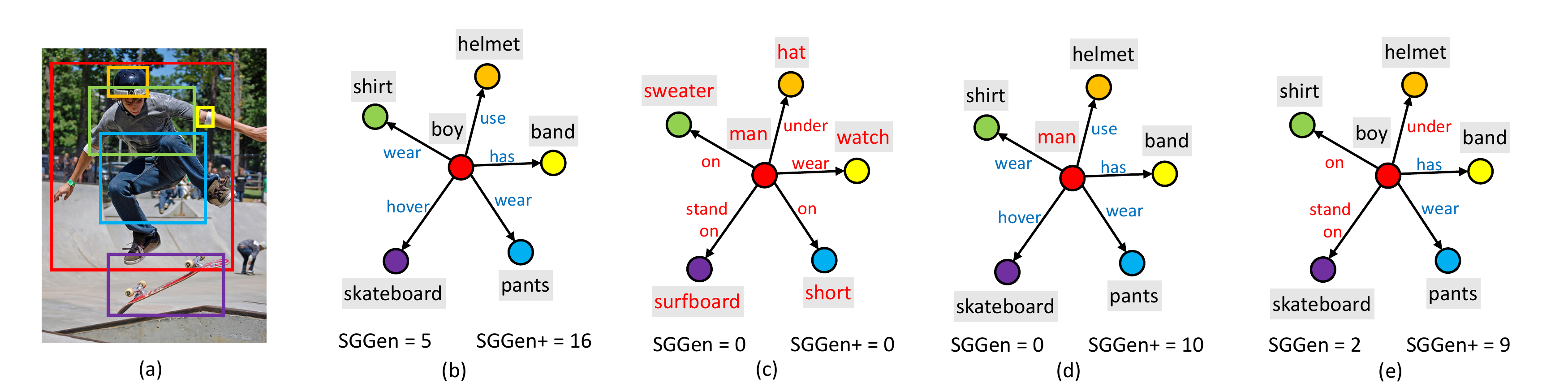}
  \end{center}
 \vspace{-3mm}
\caption{A example to demonstrate the difference between \texttt{SGGen} and \texttt{SGGen+}. Given the input image (a), its ground truth scene graph is depicted in (b). (c)-(e) are three generated scene graphs. For clarity, we merely show the connections with \textit{boy}. At the bottom of each graph, we compare the number of correct predictions for two metrics.}
\label{fig:metric}
 \vspace{-1mm}
\end{figure}

Prior work has circumvented these issues by evaluating scene graph generation under a simple triplet-recall based metric introduced in \cite{xu2017scene}. Under this metric which we will refer to as \texttt{SGGen}, the ground truth scene graph is represented as a set of $\langle \mathtt{object, relationship, subject} \rangle$ triplets and recall is computed via exact match. That is to say, a triplet is considered `matched' in a generated scene graph if all three elements have been correctly labeled, and both \texttt{object} and \texttt{subject} nodes have been properly localized (\ie, bounding box IoU $>$ 0.5). While simple to compute, this metric results in some unintuitive notions of similarity that we demonstrate in Fig.~\ref{fig:metric}.

Fig.~\ref{fig:metric}a shows an input image overlaid with bounding box localizations of correspondingly colored nodes in the ground truth scene graph shown in (b). (c), (d), and (e) present erroneously labeled scene graphs corresponding to these same localizations. Even a casual examination of (c) and (d) yields the stark difference in their accuracy -- while (d) has merely mislabeled the boy as a man, (c) has failed to accurately predict even a single node or relationship! Despite these differences, neither recalls a single complete triplet and are both scored identically under \texttt{SGGen}  (\ie, 0).

To address this issue, we propose a new metric called \texttt{SGGen+} as the augmentation of \texttt{SGGen}. \texttt{SGGen+} not only considers the triplets in the graph, but also the singletons (object and predicate). The computation of \texttt{SGGen+} can be formulated as:
\begin{equation}
Recall = \frac{C(O) + C(P) + C(T)}{N}
\end{equation}
where $C(\cdot)$ is a counting operation, and hence $C(O)$ is the number of object nodes correctly localized and recognized; $C(P)$ is for predicate. Since the location of predicate depends on the location of subject and object, only if both subject and object are correctly localized and the predicate is correctly recognized, we will count it as one. $C(T)$ is for triplet, which is the same as \texttt{SGGen}. Here, $N$ is the number of entries (the sum of number of objects, predicates and relationships) in the ground truth graph. In Fig.~\ref{fig:metric}, using our \texttt{SGGen+}, the recall for graph (c) is still 0, since all predictions are wrong. However, the recall for graph (d) is not 0 anymore since most of the object and all predicate predictions are correct, except for one wrong prediction for the red node. Based on our new metric, we can obtain a much comprehensive measurement of scene graph similarity.
\section{Experiments}
Recently, there are some inconsistencies in existing work on scene graph generation in terms of data preprocessing, data split, and evaluation. This makes it difficult to systematically benchmark progress and cleanly compare numbers across papers. So we first clarify the details of our experimental settings.

\textbf{Datasets}. There are a number of splits of the Visual Genome dataset that have been used in the scene graph generation literature \cite{xu2017scene,li2017scene,zhang2017relationship}. The most commonly used is the one proposed in \cite{xu2017scene}. Hence, in our experiments, we follow their preprocessing strategy and dataset split. After preprocessing, the dataset is split into training and test sets, which contains 75,651 images and 32,422 images, respectively. In this dataset, the top-frequent 150 object classes and 50 relation classes are selected. Each image has around 11.5 objects and 6.2 relationships in the scene graph.

\textbf{Training}. For training, multiple strategies have been used in literature. In \cite{xu2017scene,li2017scene,newell2017pixels}, the authors used two-stage training, where the object detector is pre-trained, followed by the joint training of the whole scene graph generation model. To be consistent with previous work \cite{xu2017scene,li2017scene}, 
we also adopt the two-stage training -- 
we first train the object detector and then train the whole model jointly until convergence. 

\textbf{Metrics}. We use four metrics for evaluating scene graph generation, including three previously used metrics and our proposed \texttt{SGGen+} metric:
\begin{itemize}[noitemsep]
\setlength\itemsep{0em}
\item \textbf{Predicate Classification (PredCls)}: The performance for recognizing the relation between two objects given the ground truth locations.
\item \textbf{Phrase Classification (PhrCls)}: The performance for recognizing two object categories and their relation given the ground truth locations.
\item \textbf{Scene Graph Generation (SGGen)}: The performance for detecting objects (IoU $>$ 0.5) \emph{and} recognizing the relations between object pairs.
\item \textbf{Comprehensive Scene Graph Generation (SGGen+)}: Besides the triplets counted by \texttt{SGGen}, it considers the singletons and pairs (if any), as described earlier.
\end{itemize}

\textbf{Evaluation}. 
In our experiments, we multiply the classification scores for subjects, objects and their relationships, then sort them in descending order. Based on this order, we compute the recall at top 50 and top 100, respectively. Another difference in existing literature in the evaluation protocol is w.r.t. the \texttt{PhrCls} and \texttt{PredCls} metrics. Some previous works \cite{li2017scene,newell2017pixels} used different models to evaluate along different metrics. However, such a comparison is unfair since the models could be trained to overfit the respective metrics. For meaningful evaluation, we evaluate a single model -- the one obtained after joint training -- across all metrics. 

\subsection{Implementation Details}
We use Faster R-CNN \cite{ren2015faster} associated with VGG16 \cite{simonyan2014very} as the backbone based on the PyTorch re-implementation \cite{jjfaster2rcnn}. During training, the number of proposals from RPN is 256. For each proposal, we perform ROI Align \cite{he2017mask} pooling, to get a $7 \times 7$ response map, which is then fed to a two-layer MLP to obtain each proposal's representation. In RePN, the projection functions $\Phi(\cdot)$ and $\Psi(\cdot)$ are simply two-layer MLPs. During training, we sample 128 object pairs from the quadratic number of candidates. We then obtain the union of boxes of the two objects and extract a representation for the union. The threshold for box-pair NMS is 0.7. In aGCN, to obtain the attention for one node pair, we first project the object/predicate features into 256-d and then concatenate them into 512-d, which is then fed to a two-layer MLP with a 1-d output. For aGCN, we use two aGCN layers at the feature level and semantic level, respectively. The attention on the graph is updated in each aGCN layer at the feature level, which is then fixed and sent to the aGCN at the semantic level.

\textbf{Training}. As mentioned, we perform stage-wise training -- we first pretrain Faster R-CNN for object detection, and then fix the parameters in the backbone to train the scene graph generation model. SGD is used as the optimizer, with initial learning rate 1e-2 for both training stages. 

\subsection{Analysis on New Metric}

\jianwei{We first quantitatively demonstrate the difference between our proposed metric \texttt{SGGen+} and \texttt{SGGen}. We compare them by perturbing ground truth scene graphs. We consider assigning random incorrect labels to objects; perturbing objects 1) without relationships, 2) with relationships, and 3) both. We vary the fraction of nodes which are perturbed among \{20\%, 50\%, 100\%\}. Recall is reported for both metrics. As shown in Table~\ref{Table:Exp_SGGen+}, \texttt{SGGen} is completely insensitive to the perturbation of objects without relationships (staying at 100 consistently) since it only considers relationship triplets. Note that there are on average 50.1\% objects without relationships in the dataset, which \texttt{SGGen} omits. On the other hand, \texttt{SGGen} is overly sensitive to label errors on objects with relationships (reporting 54.1 at only 20\% perturbation where the overall scene graph is still quite accurate). Note that even at 100\% perturbation the object localizations and relationships are still correct such that \texttt{SGGen+} provides a non-zero score, unlike \texttt{SGGen} which considers the graph entirely wrong. Overall, we hope this analysis demonstrates that \texttt{SCGen+} is more comprehensive compared to \texttt{SCGen}.}

\begin{table}[t]
\centering
\resizebox{0.99\columnwidth}{!}{
\setlength{\tabcolsep}{6 pt}
  \begin{tabular}{l c c c c c c c c c c c c c c}
   \toprule
   {Perturb Type} & \multicolumn{1}{c}
{{none}} & \multicolumn{3}{c}{{w/o relationship}} & \multicolumn{3}{c}{{w/ relationship}} & \multicolumn{3}{c}{{both}} \\
\cmidrule(r){2-2}
\cmidrule(r){3-5}
\cmidrule(r){6-8}
\cmidrule(r){9-11}
{Perturb Ratio} & \multicolumn{1}{c}{0\%} & \multicolumn{1}{c}{20\%}  & \multicolumn{1}{c}{50\%} & \multicolumn{1}{c}{100\%} & \multicolumn{1}{c}{20\%} & \multicolumn{1}{c}{50\%} & \multicolumn{1}{c}{100\%} & \multicolumn{1}{c}{20\%} & \multicolumn{1}{c}{50\%} & \multicolumn{1}{c}{100\%}\\
   \cmidrule(r){2-2}
   \cmidrule(r){3-4}
   \cmidrule(r){6-7}
   \cmidrule(r){8-9}
   \cmidrule(r){1-11}
   {SGGen}  & 100.0 & 100.0 & 100.0 & 100.0 & 54.1 & 22.1 & 0.0 & 62.2 & 24.2 & 0.0 \\
   {SGGen+} & 100.0 & 94.5 & 89.1 & 76.8 & 84.3 & 69.6 & 47.9 & 80.1 & 56.6 & 22.8 \\
\bottomrule
\end{tabular}}\\[3pt]
\caption{Comparisons between \texttt{SGGen} and \texttt{SGGen+} under different perturbations.}
\label{Table:Exp_SGGen+}
\end{table}

\subsection{Quantitative Comparison}
\begin{table}[t]
\footnotesize
\setlength{\tabcolsep}{2.5pt} 
\center
\resizebox{0.95\columnwidth}{!}{
  \begin{tabular}{c l c c c c c c c c}
   \toprule
  & \multicolumn{1}{c}{} & \multicolumn{2}{c}{SGGen+}  & \multicolumn{2}{c}{SGGen} & \multicolumn{2}{c}{PhrCls}  & \multicolumn{2}{c}{PredCls} \\    
   \cmidrule(r){3-4}
   \cmidrule(r){5-6}
   \cmidrule(r){7-8}
   \cmidrule(r){9-10}
	& Method & R@50 & R@100 & R@50 & R@100 & R@50 & R@100 & R@50 & R@100\\
    \cmidrule(r){2-10}
	& IMP \cite{xu2017scene} & - & - & 3.4 & 4.2 & 21.7 & 24.4 & 44.8 & 53.0 \\
    & MSDN \cite{li2017scene} & - & -  & 7.7 & 10.5 & 19.3 & 21.8 & 63.1 & 66.4 \\
    & Pixel2Graph \cite{newell2017pixels} & - & - & 9.7 & 11.3 & 26.5 & 30.0 & 68.0 & 75.2 \\
    \cmidrule(r){2-10}
    & IMP$^\dagger$ \cite{xu2017scene} & 25.6 & 27.7 & 6.4 & 8.0 & 20.6 & 22.4 & 40.8 & 45.2 \\
    & MSDN$^\dagger$ \cite{li2017scene} & 25.8 & 28.2  & 7.0 & 9.1 & 27.6 & 29.9 & 53.2 & 57.9 \\ 
    & NM-Freq$^\dagger$ \cite{zellers2017neural} & 26.4 & 27.8  & 6.9 & 9.1 & 23.8 & 27.2 & 41.8 & 48.8 \\       
	& Graph R-CNN (Us) & \textbf{28.5} & \textbf{35.9} & \textbf{11.4} & \textbf{13.7} & \textbf{29.6} & \textbf{31.6} & \textbf{54.2} & \textbf{59.1} \\ 
\bottomrule
\end{tabular}}
\caption{Comparison on Visual Genome test set \cite{krishna2017visual}. We reimplemented IMP \cite{xu2017scene} and MSDN \cite{li2017scene} using the same object detection backbone for fair comparison.}
\label{Table:Exp_main}
\end{table}

We compare our Graph R-CNN with recent proposed methods, including Iterative Message Passing (IMP) ~\cite{xu2017scene}, Multi-level scene Description Network (MSDN)~\cite{li2017scene}. Furthermore, we evaluate the neural motif frequency baseline proposed in \cite{zellers2017neural}. Note that previous methods often use slightly different pre-training procedures or data split or extra supervisions. For a fair comparison and to control for such orthogonal variations, we reimplemented IMP, MSDN and frequency baseline in our codebase. Then, we re-train IMP and MSDN based on our backbone -- specifically, we used the same pre-trained object detector, and then jointly train the scene graph generator until convergence. We denote these as IMP$^\dagger$ and MSDN$^\dagger$. Using the same pre-trained object detector, we report the neural motif frequency baseline in \cite{zellers2017neural} as NM-Freq$^\dagger$.

We report the scene graph generation performance in Table~\ref{Table:Exp_main}. The top three rows are numbers reported in the original paper, and the bottom four rows are the numbers from our re-implementations. First, we note that our re-implementations of IMP and MSDN (IMP$^\dagger$ and MSDN$^\dagger$) result in performance that is close to or better than the originally reported numbers under some metrics (but not all), which establishes that the takeaway messages next are indeed due to our proposed architectural choices -- relation proposal network and attentional GCNs. Next, we notice that Graph R-CNN outperforms IMP$^\dagger$ and MSDN$^\dagger$. This indicates that our proposed Graph R-CNN model is more effective to extract the scene graph from images. Our approach also outperforms the frequency baseline on all metrics, demonstrating that our model has not just learned simple co-occurrence statistics from training data, but rather also captures context in individual images. More comprehensively, we compare with IMP and MSDN on the efficiency over training and inference. IMP uses 2.15$\times$ while MSDN uses 1.86$\times$ our method. During inference, IMP is 3.27$\times$ while MSDN is 3.80$\times$ slower than our Graph R-CNN. This is mainly due to the simplified architecture design (especially the aGCN for context propagation) in our model.

\subsection{Ablation Study}
\begin{figure}[t]
\begin{center}
\includegraphics[ width=1.0\textwidth]{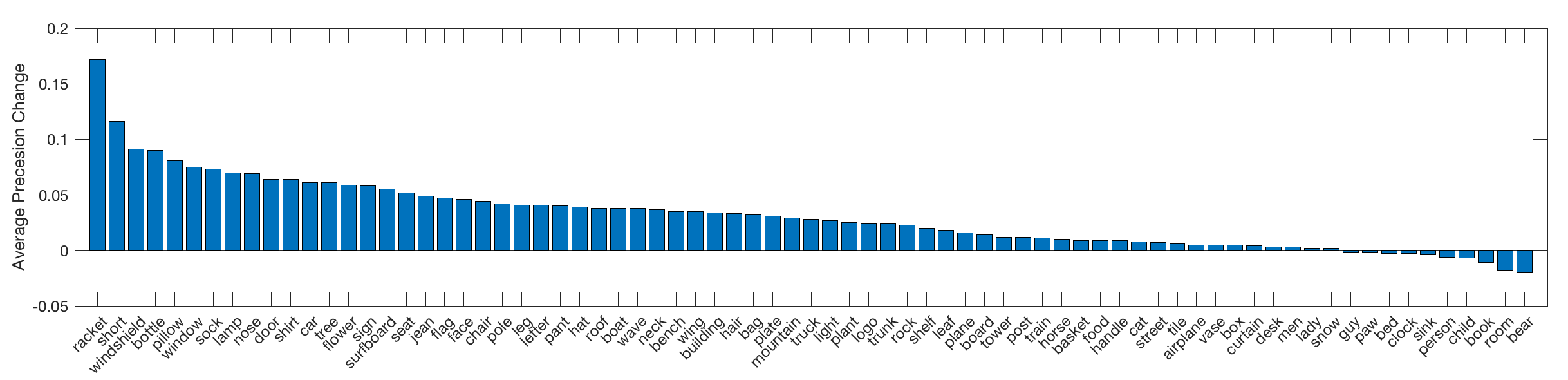}
\end{center}%
\caption{Per category object detection performance change after adding RePN.}
\label{fig:ap}
\end{figure}

\begin{table}[t]\footnotesize
\setlength{\tabcolsep}{1.8pt}
\centering
\resizebox{0.95\columnwidth}{!}{
\begin{tabular}{c c c c ccccccccc}
\toprule
\multirow{3}{*}{RePN} & \multirow{3}{*}{GCN} & \multirow{3}{*}{aGCN} & {Detection} & \multicolumn{2}{c}{SGGen+} & \multicolumn{2}{c}{SGGen} & \multicolumn{2}{c}{PhrCls} & \multicolumn{2}{c}{PredCls}\\
\cmidrule(r){4-4}
\cmidrule(r){5-6} 
\cmidrule(r){7-8}
\cmidrule(r){9-10}
\cmidrule(r){11-12}
& & & mAP@0.5 & R@50 & R@100 & R@50 & R@100 & R@50 & R@100 & R@50 & R@100 \\
\midrule
 - & - & - &  20.4    &  25.9 & 27.9 & 6.1 & 7.9 & 17.8  & 19.9 & 33.5 & 38.4 \\
\checkmark & - & - &  \textbf{23.6} & 27.6 & 34.8 &  8.7 & 11.1  & 18.3  &  20.4 & 34.5 & 39.5 \\
\checkmark & \checkmark & - & 23.4 & 28.1 & 35.3 & 10.8 & 13.4 & 27.2  & 29.5 & 52.3 & 57.2 \\
\checkmark & - & \checkmark & 23.0 & \textbf{28.5} & \textbf{35.9} & \textbf{11.4} & \textbf{13.7} & \textbf{29.4} & \textbf{31.6} & \textbf{54.2} & \textbf{59.1}    \\
\bottomrule
\end{tabular}}\\[5pt]
\caption{Ablation studies on Graph R-CNN. We report the performance based on four scene graph generation metrics and the object detection performance in mAP@0.5.}
\label{Table:Exp_Ablation}
\end{table}

In Graph R-CNN, we proposed two novel modules -- relation proposal network (RePN) and attentional GCNs (aGCN). In this sub-section, we perform ablation studies to get a clear sense of how these different components affect the final performance. The left-most columns in Table~\ref{Table:Exp_Ablation} indicate whether or not we used RePN, GCN, and attentional GCN (aGCN) in our approach. 
The results are reported in the remaining columns of Table~\ref{Table:Exp_Ablation}. We also report object detection performance mAP@0.5 following Pascal VOC's metric \cite{everingham2012pascal}.

In Table~\ref{Table:Exp_Ablation}, we find RePN boosts \texttt{SGGen} and \texttt{SGGen+} significantly. This indicates that our RePN can effectively prune the spurious connections between objects to achieve high recall for the correct relationships. We also notice it improves object detection significantly. In Fig.~\ref{fig:ap} we show the per category object detection performance change when RePN is added. For visual clarity, we dropped every other column when producing the plot. We can see that almost all object categories improve after adding RePN. Interestingly, we find the detection performance on categories like \textit{racket}, \textit{short}, \textit{windshield}, \textit{bottle} are most significantly improved. 
Note that many of these classes are smaller objects that have strong relationships with other objects, e.g.~ rackets are often carried by people. Evaluating \texttt{PhrCls} and \texttt{PredCls} involves using the ground truth object locations. Since the number of objects in images (typically $<$25) is much less than the number of object proposals (64), the number of relation pairs is already very small. As a result, RePN has less effect on these two metrics.

\begin{figure}[t]
  \begin{center}
    \includegraphics[width=0.95\textwidth, trim=0 1.2cm 1.2cm 0, clip]{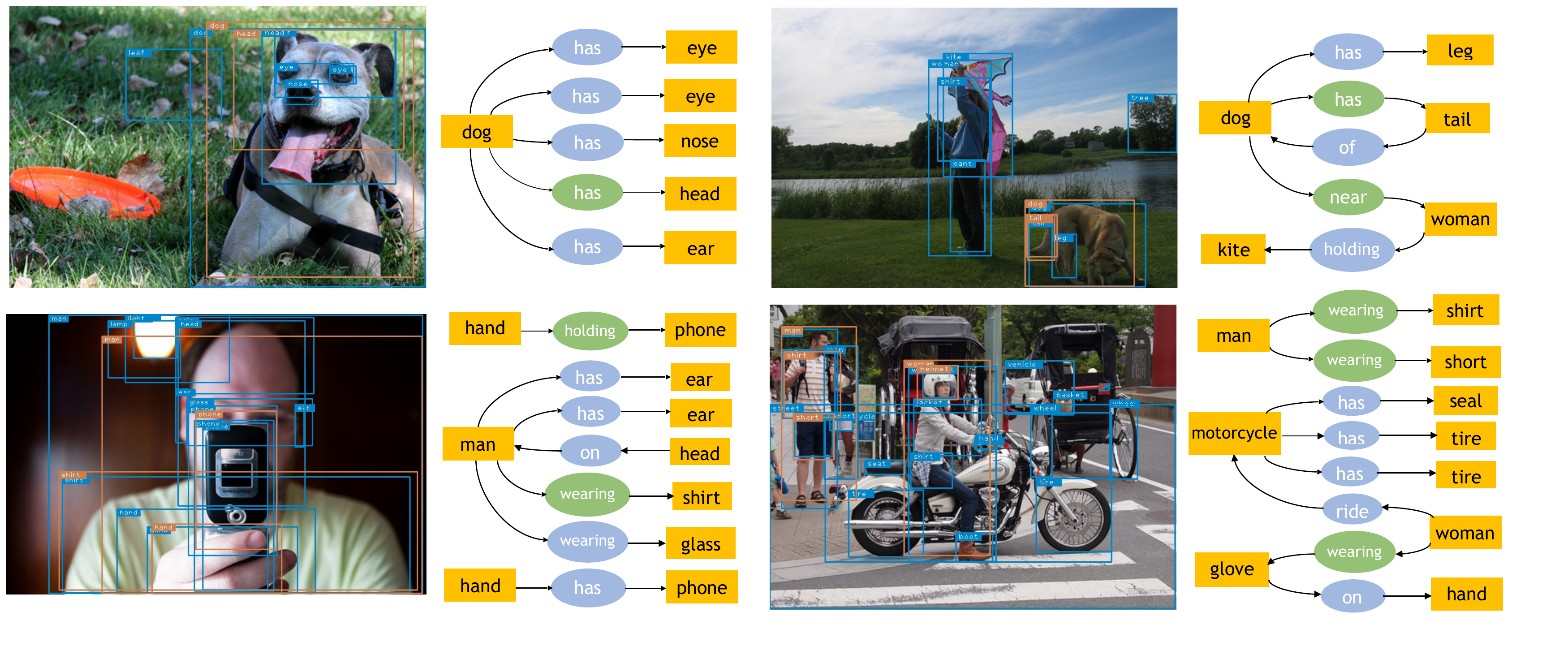}
  \end{center}
\caption{Qualitative results from Graph R-CNN. In images, blue and orange bounding boxes are ground truths and correct predictions, respectively. In scene graphs, blue ellipsoids are ground truth relationships while green ones denote correct predictions.}
\label{fig:qualitative}
\end{figure}

By adding the aGCNs into our model, the performance is further improved. These improvements demonstrate that the aGCN in our Graph R-CNN can capture meaningful context across the graph. We also compare the performance of our model with and without attention. We see that by adding attention on top of GCNs, the performance is higher. This indicates that controlling the extent to which contextual information flows through the edges is important. These results align with our intuitions mentioned in the introduction. Fig.~\ref{fig:qualitative} shows generated scene graphs for test images. With RePN and aGCN, our model is able to generate higher recall scene graphs. The green ellipsoids shows the correct relationship predictions in the generated scene graph.

\section{Conclusion}

In this work, we introduce a new model for scene graph generation -- Graph R-CNN. Our model includes a relation proposal network (RePN) that efficiently and intelligently prunes out pairs of objects that are unlikely to be related, and an attentional graph convolutational network (aGCN) that effectively propagates contextual information across the graph. We also introdce a novel scene graph generation evaluation metric (\texttt{SGGen+}) that is more fine-grained and realistic than existing metrics. Our approach outperforms existing methods for scene graph generation, as evaluated using existing metrics and our proposed metric.

\noindent\small{\textbf{Acknowledgements}.} This work was supported in part by NSF, AFRL, DARPA, Siemens, Google, Amazon, ONR YIPs and ONR Grants N00014-16-1-\{2713,2793\}.

%
%
%
\bibliographystyle{splncs04}
\bibliography{mybib}

\end{document}